\documentclass[10pt,twocolumn,letterpaper]{article}

 \usepackage[pagenumbers]{cvpr} 

\usepackage[dvipsnames]{xcolor}

\usepackage{amsmath}
\usepackage{amssymb}
\usepackage{mathtools}
\usepackage{amsthm}
\usepackage{amsfonts,mathrsfs}

\numberwithin{equation}{section}

\theoremstyle{plain}
\newtheorem{theorem}{Theorem}[section]

\theoremstyle{definition}
\newtheorem{definition}[theorem]{Definition}

\theoremstyle{remark}
\newtheorem{remark}[theorem]{Remark}

\newcommand\R{\mathbb{R}}

\definecolor{cvprblue}{rgb}{0.21,0.49,0.74}
\usepackage[pagebackref,breaklinks,colorlinks,citecolor=cvprblue]{hyperref}

\def\paperID{14112} 
\def\confName{CVPR}
\def\confYear{2024}

\title{From Activation to Initialization: Scaling Insights for Optimizing Neural Fields}

\author{Hemanth Saratchandran$^{*}$\\
Australian Institute of Machine Learning,\\
University of Adelaide, Australia\\
\and
Sameera Ramasinghe\\
Amazon, Australia\\
\and
Simon Lucey\\
Australian Institute of Machine Learning,\\
University of Adelaide, Australia\\
}

\begin{document}
\maketitle
\def\thefootnote{*}\footnotetext{hemanth.saratchandran@adelaide.edu.au}\def\thefootnote{\arabic{footnote}}
\begin{abstract}
In the realm of computer vision, Neural Fields have gained prominence as a contemporary tool harnessing neural networks for signal representation. Despite the remarkable progress in adapting these networks to solve a variety of problems, the field still lacks a comprehensive theoretical framework. This article aims to address this gap by delving into the intricate interplay between initialization and activation, providing a foundational basis for the robust optimization of Neural Fields. Our theoretical insights reveal a deep-seated connection among network initialization, architectural choices, and the optimization process, emphasizing the need for a holistic approach when designing cutting-edge Neural Fields. 
\end{abstract}    
\section{Introduction}
\label{sec:intro}

Neural Fields have emerged as a compelling paradigm leveraging coordinate-based neural networks to achieve a concise and expressive encoding of intricate geometric structures and visual phenomena 
\cite{nerf, chen2021learning, chen2022fully, li20223d, sitzmann2019scene, tancik2020fourier}. 
Despite the notable advancements in the application of neural fields across various domains \cite{chng2022garf, skorokhodov2021adversarial, sun2021coil, ramasinghe2023effectiveness}, the prevalent approach to understanding and designing these networks remains primarily empirical. Additionally, in the era of big data, practitioners often follow the trend of scaling networks ad hoc with larger datasets, lacking a clear understanding of how such architectures should proportionately adapt to data size. 


In this paper, we explore the scaling dynamics of neural fields in relation to data size. Specifically, when given a neural field and a dataset, we inquire about the number of parameters necessary for the neural architecture to facilitate gradient descent convergence to a global minimum. 
Our theoretical findings imply that the answer to this question depends on the chosen activation and initialization scheme.  For shallow networks employing a sine \citep{sitzmann2020implicit}, sinc \cite{ramasinghe2023effectiveness}, Gaussian \cite{ramasinghe22}, or wavelet activation \cite{saragadam2023wire} and initialized with standard schemes such as LeCun \cite{lecun2002efficient}, Kaiming \cite{he2015delving}, or Xavier \cite{glorot2010understanding}, the network parameters must scale super-linearly with the number of training samples for gradient descent to converge effectively. In the case of deep networks with the same activations and initializations, we prove that the network parameters need to scale super-quadratically. This contrasts with the work \cite{nguyen2021proof}, demonstrating that networks employing a ReLU activation, with or without a positional embedding layer \cite{tancik2020fourier}, scale quadratically in the shallow setting and cubically in the deep setting.  Other studies have explored analogous scaling laws \cite{agustsson2017ntire, allen2019convergence, allen2019learning, arora2019fine, oymak2020toward}, yet in each instance, the scaling demands were cubic or more and pertained to initializations not commonly employed by practitioners in the field. While Nguyen's work \cite{nguyen2021proof} was previously considered state-of-the-art, our theoretical insights challenge this notion, demonstrating that significantly fewer parameters are required when employing a different activation function. For further comparison of our results with the literature we refer the reader to the related work sec. \ref{sec;rel_work}.

\begin{figure}[t]
    \centering
    \includegraphics[width=0.8\linewidth]
    {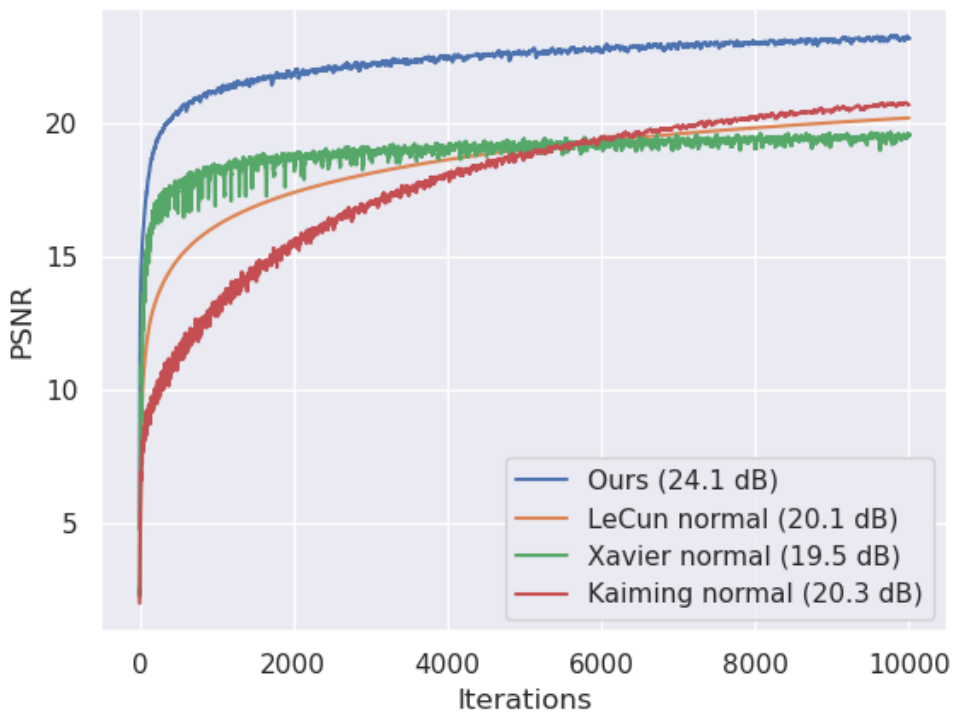}
    \caption{We evaluate Gaussian-activated networks comprising four hidden layers, initialized using four different methods, and trained with full-batch gradient descent. The comparison is performed on an image reconstruction task, with the final train PSNRs displayed in parentheses in the legend.}
    \label{fig:front_fig}
\end{figure}

Theoretical results often find themselves in regimes that may not align with practical applications, rendering the theory insightful but lacking in predictiveness. 
To underscore the predictive efficacy of our theoretical framework, we design a novel initialization scheme and demonstrate its superior optimality compared to standard practices in the literature. Our findings reveal that neural fields, equipped with the activations sine, sinc, Gaussian, or wavelet and initialized using our proposed scheme, require a linear scaling with data in the shallow case and a quadratic scaling in the deep case, for gradient descent to converge to a global optimum. When compared with standard practical initializations such as LeCun \cite{lecun2002efficient}, Xavier \cite{glorot2010understanding}, and Kaiming \cite{he2015delving}, our initialization proves to be significantly more parameter-efficient. We turn the readers attention to fig. \ref{fig:front_fig} where we compared our initialization with Kaiming normal \cite{he2015delving}, Xavier normal \cite{glorot2010understanding} and Lecun Normal \cite{lecun2002efficient} on an image reconstruction task. As predicted by our theory, our initialization shows superior performance.



We extensively test our initialization scheme across various neural field applications including image regression, super resolution, shape reconstruction, tomographic reconstruction, novel view synthesis and physical modeling. Our contributions include:
\begin{itemize}
    \item[1.] The first theoretical proof of scaling laws for neural fields with the activations sine, sinc, Gaussian, and wavelets, ensuring effective convergence with gradient descent. The proof demonstrates that networks employing these activations surpass state-of-the-art outcomes in terms of parameter efficiency.

    \item[2.] The development of a superior initialization scheme compared to standard approaches in the literature.

    \item[3.] Empirical validation of our theoretical predictions on various neural field applications
\end{itemize}

\section{Related Work}\label{sec;rel_work}
Several works have considered the effect of overparameterization on gradient descent convergence.
The work \cite{du2019gradient} considered convergence of gradient decent for smooth 
activations using the Neural Tangent Kernel (NTK) parameterization \cite{jacot2018neural} and showed that if 
all hidden layers satisfied the growth $\Omega(N^4)$, then gradient decent converges 
to a global minimum. 
In \cite{huang2020dynamics}, using the Neural Tangent Hierarchy 
it was shown that for a smooth activation, $\Omega(N^3)$ suffices for all the hidden 
layers to guarantee convergence of gradient decent to a global minimum.  Both these papers used the standard normal distribution as initialization $\mathcal{N}(0,1)$, which is rarely used in practise especially in neural fields applications.
There have been several works that have studied the convergence of gradient decent for 
ReLU activated neural networks, \cite{allen2019convergence, zou2019improved} prove 
convergence for gradient decent in the setting where the input and output layers are 
fixed, while the inner layers are only trained. Their result requires polynomial overparameterization for a large degree polynomial.
For two layer ReLU networks, the works 
\cite{arora2019fine, du2019gradient, oymak2020toward} study the convergence of gradient decent essentially showing that the width scaling must be at the order of
$\Omega(N^4)$. For deep ReLU networks, the state of the art was proved in 
\cite{nguyen2021proof} showing that one could take one hidden layer of order
$\Omega(N^3)$. In each case our results show that much less overparamterization is needed and our analysis is carried out with initializations used by practitioners.

\section{Notation}\label{sec;notation}

We consider a depth $L$ neural network with layer widths 
$\{n_1,\ldots,n_L\}$. We let $X \in \R^{N\times n_0}$ denote the training data, with $n_0$ being the dimension of the input and $N$ being the number of training samples. We let 
$Y \in \R^{N\times n_L}$ denote the training labels. The output at layer $k$ will be denoted by $F_k$ and is defined by
\begin{equation}\label{defn_net}
    F_k =  
    \begin{cases}
        F_{L-1}W_L + b_L, & k = L \\
        \phi(F_{k-1}W_k + b_k), & k \in [L-1] \\
        X, & k = 0
    \end{cases}
\end{equation}
where the weights $W_k \in \R^{n_{k-1}\times n_k}$ and the biases 
$b_k \in \R^{n_k}$ and $\phi$ is an activation applied component wise.
The notation
$[m]$ is defined by $[m] = \{1,\ldots,m\}$.  For a weight matrix $W_k$ at layer k, the notation $W_k^0$ will denote the initialization of that weight matrix. These are the initial weights of the network before training begins under a gradient descent algorithm. The networks we use in the paper will all be trained using the MSE loss function, which we denote by
$\mathcal{L}$.
Our theoretical results will be primarily for the case where $\phi$ is given by one of the activation 
sine, sinc, Gaussian or wavelet. We remind the reader that the sinc function is defined by: $sinc(x) = \frac{sin(x)}{x}$ for $x \neq 0$ and $sinc(0) = 1$. For more details on how these activations are used within the context of neural fields we refer the reader to \cite{siren, ramasinghe22, saragadam2023wire, ramasinghe2023effectiveness}.

We will always assume the data set $X$ consists of i.i.d sub-gaussian vectors and for the theoretical proofs will assume they are normalized to have norm $\vert\vert X_i\vert\vert_2 = 1$.
More details on data assumptions can be found in sec. 1 of the supp. material. 

Our networks will be free to have a positional embedding layer (PE), which is simply an embedding of the data into a higher dimensional space. The reader who is unfamiliar with PE should consult the standard references \cite{tancik2020fourier, nerf}.

A neural field is any such network that parameterizes a continuous field. Examples of neural fields can be found in sec. \ref{sec;exps}. All networks will be trained with the standard Mean Square Error (MSE) loss.


We will use standard complexity notations, $\Omega(\cdot)$, 
$\mathcal{O}(\cdot)$, $\Theta(\cdot)$ throughout the paper. The reader who is unfamiliar with this notation can consult sec. 1 of the supp. material. Finally, we will use the notation "w.h.p." to denote \textit{with high probability}.



\section{Theoretical Scaling Laws}\label{sec;theory_main}

In this section, we provide a theoretical understanding of how much overparameterization a neural field needs in order for gradient descent to converge to a global minimum for a given data set. Furthermore, we prove a scaling law that details how the network must scale as the dataset size grows, in order to facilitate optimum convergence of gradient descent. We will work with the activations sine, sinc and Gaussian. For wavelets see sec. 1 of the supp material.
To accommodate space constraints and offer readers a more expansive perspective, we have deferred the exhaustive details of the proofs to sec. 1 of the supp. material.

We begin with a definition of overparameterization that we will use throughout the paper. Our definition is consistent with several works in the literature \cite{cooper2018loss, zou2019improved, nguyen2021proof, allen2019learning}.

\begin{definition}
    Given a data set $X$ with $N$ samples and a neural network
    $F(\theta)$. We say the network $F(\theta)$ is overparameterized with respect to the size of $X$ if the number of parameters $\theta$ is greater than $N$.
\end{definition}

In general, it is known via several works in the literature that under various assumptions, overparameterization is necessary for gradient descent to converge to a global minimum \cite{allen2019convergence, agustsson2017ntire, arora2019fine, du2019gradient, oymak2020toward, nguyen2021proof}

\subsection{A scaling law for shallow  networks}\label{sec;shallow_nets}

In this section, we present our theorem, delineating a complexity bound that dictates the extent of overparameterization necessary for a neural field to enable gradient descent convergence to a global minimum, particularly in the context of shallow networks—those with only two layers. 


\begin{theorem}\label{thm;Lecun_shallow}
Let $X$ be a fixed data set with $N$ samples.
Let $F$ be a shallow neural network of depth $2$ admitting one of the following activation functions:
\begin{itemize}
    \item[1.] $sin(\omega x)$,
    \item[2.] $sinc(\omega x) = \frac{sin(\omega x)}{x}$,
    \item[3.] $e^{-x^2/2\omega^2}$
\end{itemize}
where $\omega$ ($1/\omega^2$) is a fixed frequency hyperparameter.
Let the widths of the network satisfy
\begin{align}
    n_1 = \Omega(N^{3/2}) \text{ and }
    n_2 = \Theta(m) 
    \end{align}
where $m$ is a fixed positive constant.
Suppose the network has been initialized according to LeCun's initialization scheme
\begin{equation}\label{eqn;lecun_init_shallow}
    (W^0_{1})_{ij} \sim 
    \mathcal{N}(0, \frac{1}{n_0}) \text{ and }
    (W^0_{2})_{ij} \sim \mathcal{N}(0, \frac{1}{n_1}).
\end{equation}
Then for a small enough learning rate gradient descent converges to a global minimum w.h.p. when trained on $X$.
\end{theorem}

\begin{remark}\label{rmk;kaimingshallow}
    Thm. \ref{thm;Lecun_shallow} has been expressed in the context of LeCun's initialization. However, a similar theorem can be proved for the two other standard initializations used in the literature, namely Xavier normal initialization \cite{glorot2010understanding} and Kaiming normal initialization \cite{he2015delving}.
\end{remark}


\subsection{A scaling law for deep networks}\label{sec;deep_nets}

In this section we present a generalization of thm. \ref{thm;Lecun_shallow} to the setting of deep networks.

\begin{theorem}\label{thm;lecun_deep}
Let $X$ be a fixed data set of size $N$.
    Let $F$ be a deep neural network of depth $L$, $L > 2$, admitting one of the following activation functions:
\begin{itemize}
    \item[1.] $sin(\omega x)$,
    \item[2.] $sinc(\omega x) = \frac{sin(\omega x)}{x}$,
    \item[3.] $e^{-x^2/2\omega^2}$
\end{itemize}
where $\omega > 0$ ($1/\omega^2$) is a fixed frequency hyperparameter.
Let the widths of the network satisfy
\begin{align}
    n_l &= \Theta(m)\text{, } \forall l \in [L] \text{ with } l\neq L-1 \text{, } \label{width_cond1_lecun} \\
    n_{L-1} &= \Omega(N^{5/2}), \label{width_cond2_lecun}
\end{align}
where $m$ is a fixed constant that is allowed to depend linearly on $N$. Suppose the network is initialized according to LeCun's initialization scheme
\begin{equation}\label{eqn;lecun_init_deep}
    (W^0_l)_{ij} \sim \mathcal{N}(0, 1/n_{l-1})\text{, for } l \in [L].
\end{equation}
Then for a small enough learning rate gradient descent converges to a global minimum w.h.p. when trained on $X$.
\end{theorem}

\begin{remark}\label{rmk;deep_net_compare}
  A similar theorem can be proved for Xavier and Kaiming normal initializations. See sec. 1 of the supp. material for learning rate bounds.
\end{remark}

\begin{remark}\label{rmk;connection_act_init}
Thms. \ref{thm;Lecun_shallow} and \ref{thm;lecun_deep} reveal a significant connection between activation functions and initialization methods in neural network training. These theorems underscore the importance of carefully choosing the architecture (via activation functions) and initialization for crafting parameter-efficient networks that converge to optimal solutions. 
\end{remark}



\subsection{Analyzing the proof methodology}\label{sec;proof_method}

In sec. \ref{sec;shallow_nets} and \ref{sec;deep_nets}, we explored the impact of initialization and activation choices on overparameterization in order for gradient descent to converge to a global minimum. This naturally leads to the question of whether there are initialization schemes for neural fields utilizing the activations $sin(\omega x)$, $sinc(\omega x)$, and $e^{-x^2/2\omega^2}$ that necessitate less overparameterization compared to the results established in thms. \ref{thm;Lecun_shallow} and \ref{thm;lecun_deep}.

In order to answer this question we start by giving a concise sketch of the main idea of the proof methodology of thm. \ref{thm;Lecun_shallow}. We let $F$ be a fixed 2-layer network.
Let $\sigma_0 = \sigma_{\min}(F_1^0)$, denote the smallest singular value of the hidden layer of $F$ and let 
$W_2^0 \in \R^{n_1 \times n_2}$ denote the initial weight matrix of the second layer of $F$.
The key step in proving thm. \ref{thm;Lecun_shallow} is to establish the lower bound
\begin{equation}\label{eqn;first_proof_lecun_1}
    \sigma_0^2 \geq 16\sqrt{N}\sqrt{n_1}
    \sqrt{2\mathcal{L}(\theta_0)}\vert\vert W_2^0\vert\vert_2
\end{equation}
when $n_1 = \Omega( \lceil N^{3/2} \rceil)$. Once this is achieved a routine use of convergence theory leads to the proof of thm. \ref{thm;Lecun_shallow}. The interested reader can consult sec. 1 of supp. material for full details.

To obtain \eqref{eqn;first_proof_lecun_1}, the proof proceeds by first establishing the following two complexity bounds:
\begin{align}
      \sigma_0 \geq \Omega(\sqrt{n_1}) \text{ and }
      \sqrt{2\mathcal{L}(\theta_0)} = 
      \mathcal{O}(\sqrt{N}). \label{eqn;sigma_0 and loss}
\end{align}
\textit{We pause to mention that the two inequalities in 
\eqref{eqn;sigma_0 and loss} are activation dependent and is precisely the place where we need to use that the activation is one of sine, Gaussian, sinc, or wavelet.}

Substituting \eqref{eqn;sigma_0 and loss} into \eqref{eqn;first_proof_lecun_1} shows us that
proving inequality \eqref{eqn;first_proof_lecun_1} boils down to showing 
\begin{equation}\label{eqn;first_proof_lecun_2}
    n_1 \geq CNn_1^{3/4}\vert\vert W_2^0\vert\vert_2,
\end{equation}
    where $C > 0$ is a constant (coming from the complexity bounds in \eqref{eqn;sigma_0 and loss}) which we won't worry about for this discussion. In order to establish inequality 
    \eqref{eqn;first_proof_lecun_2} we need to understand the 2-norm of the random matrix $W_2^0$. This is precisely where LeCun's initialization is used. By appealing to thm. 2.13 of \cite{davidson2001local}, it can be shown that if a random $n_1 \times n_2$-matrix $W$ has entries sampled from a Normal distribution of the form $\mathcal{N}(0, 1/n_1)$ then
    \begin{equation}\label{eqn;DS_general}
        \vert\vert W\vert\vert_2 = 
        \mathcal{O}(\sqrt{n_2}/\sqrt{n_1}).
    \end{equation}
Applying \eqref{eqn;DS_general} to the random weight matrix 
$W_2^0$, initialized using LeCun's initialization \eqref{eqn;lecun_init_shallow}, we see that \eqref{eqn;first_proof_lecun_2} holds if 
$n_1 = \Omega(N^{3/2})$. 

Inequality \eqref{eqn;first_proof_lecun_2} illuminates the intrinsic connection between initialization and overparameterization. The reduction of the product $n_1^{3/4}\vert W_2^0\vert_2$ serves as a theoretical indicator for the diminished necessity of overparameterization. Achieving a small value for $n_1^{3/4}\vert W_2^0\vert_2$ entails minimizing the norm $\vert W_2^0\vert_2$, and leveraging \eqref{eqn;DS_general} (refer to thm. 2.13 in \cite{davidson2001local}), we ascertain that the norm $\vert W_2^0\vert_2$ possesses a complexity bound of $\mathcal{O}(\sqrt{n_2}/\sqrt{n_1})$. Consequently, for a smaller norm 
$\vert W_2^0\vert_2$, theoretical sampling from a Gaussian distribution with an exceedingly small variance is warranted. Suppose the entries of $W_2^0$ are sampled from $\mathcal{N}(0, 1/n_1^p)$. Employing \eqref{eqn;DS_general} once more, we deduce that $\vert W_2^0\vert_2 = \mathcal{O}(\sqrt{n_2}/n_1^{p/2})$. Substituting this result back into \eqref{eqn;first_proof_lecun_2}, we observe that a larger value of $p$ corresponds to a reduced complexity requirement for $n_1$. In essence, \textbf{\textit{sampling the final layers weight matrix from a Normal distribution with smaller variance necessitates less overparameterization for gradient descent to converge to a global minimum}}. However, it's crucial to note that excessively small variances in the Gaussian distribution pose the challenge of vanishing gradients in the network.

While the preceding discussion focused on shallow networks, a parallel argument can be extended to deep networks, illustrating that the level of overparameterization required for the last hidden layer to achieve convergence to a global minimum is linked to the variance of the Normal distribution from which we initialize the final layer's weights. Moreover, a smaller variance corresponds to a reduced need for overparameterization. 

\subsection{Designing new initializations}\label{sec;new_inits}

The discussion in the previous section suggested a new approach to initializing weights for a neural network. The main point was that by controlling the variance of the Normal distribution the final layer weights are sampled from, we can use less parameters and still converge to a global minimum under gradient descent.

\begin{figure}[t]
    \centering
    \includegraphics[width=1.\linewidth]
    {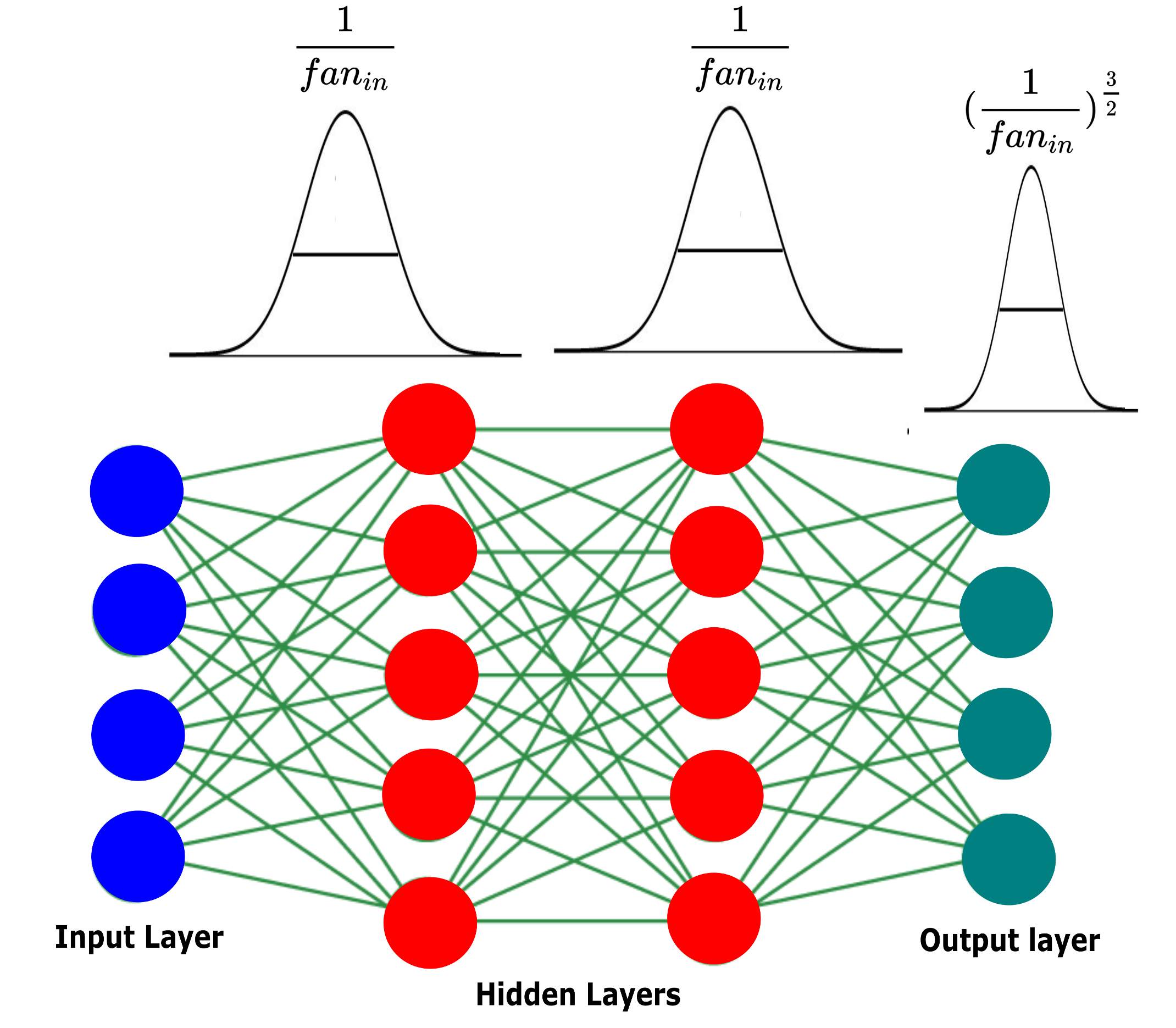}
    \caption{Diagram showing how to initialize weight matrices according to \textbf{Initialization 1}. The final output layer is initialized with a Normal distribution of smaller variance than the previous layers by a factor of 
    $1/\sqrt{fan_{in}}$, where $fan_{in}$ denotes the input dimension to the layer.}
    \label{fig:network_inits}
\end{figure}

Fix a deep neural network $F$ with $L$ layers. We define the following initialization. 
\begin{align}
    &\text{ \textbf{Initialization 1:} }  \nonumber \\
    &(W^0_l)_{ij} \sim \mathcal{N}(0, 1/n_{l-1})\text{ for } l \in [L-1]. \label{eqn;my_init_deep_a} \\
    &(W^0_L)_{ij} \sim \mathcal{N}(0, 2/(n_{l-1}^{3/2})). \label{eqn;my_init_deep_b}
\end{align}
As suggested by the discussion in sec. \ref{sec;proof_method} the variance of the Gaussian we sample from for the last layer is smaller by a factor of $1/\sqrt{n_{L-1}}$.
The biases of the network will be initialized to $0$ or a very small number such as $0.01$ as is the standard practice for many common normal initializations. Fig. \ref{fig:network_inits}, gives a diagrammatic rendition of initialization 1.



\begin{theorem}\label{thm;my_init_shallow_main}
Let $X$ be a fixed dataset with $N$ training samples.
Let $F$ be a shallow neural network of depth $2$ admitting one of the following activation functions:
\begin{itemize}
    \item[1.] $sin(\omega x)$,
    \item[2.] $sinc(\omega x) = \frac{sin(\omega x)}{x}$,
    \item[3.] $e^{-x^2/2\omega^2}$. 
\end{itemize}
Let the widths of the network satisfy
\begin{align}
    n_1 = \Omega(N) \text{ and }
    n_2 = \Theta(m) 
    \end{align}
where $m$ is a fixed positive constant.
Suppose the network has been initialized according to initialization 1, see \eqref{eqn;my_init_deep_a}, \eqref{eqn;my_init_deep_b}.
Then for a small enough learning rate gradient descent converges to a global minimum w.h.p.
\end{theorem}

\begin{remark}\label{rmk;my_shallow_vs_lecun_shallow}
Comparing thm. \ref{thm;my_init_shallow_main} with thm. \ref{thm;Lecun_shallow}, we observe that utilizing Initialization 1 requires only linear overparameterization in the final hidden layer, as opposed to superlinear overparameterization.
\end{remark}

\begin{theorem}\label{thm;my_init_deep_main}
    Let $X$ be a fixed dataset with $N$ training samples. Let $F$ be a deep neural network of depth $L$, $L > 2$, admitting one of the following activation functions:
\begin{itemize}
    \item[1.] $sin(\omega x)$,
    \item[2.] $sinc(\omega x) = \frac{sin(\omega x)}{x}$,
    \item[3.] $e^{-x^2/2\omega^2}$. 
\end{itemize}
Let the widths of the network satisfy
\begin{align}
    n_l &= \Theta(m)\text{, } \forall l \in [L] \text{ with } l\neq L-1 \text{, } \label{width_cond1_mine} \\
    n_{L-1} &= \Omega(N^{2}), \label{width_cond2_mine} 
\end{align}
where $m$ is a fixed constant that is allowed to depend linearly on $N$. Suppose the network is initialized by initialization 1, 
see \eqref{eqn;my_init_deep_a}, \eqref{eqn;my_init_deep_b}.
Then for a small enough learning rate gradient descent converges to a global minimum w.h.p.
\end{theorem}

\begin{remark}\label{rmk;my_deep_vs_lecun_deep}
Comparing thm. \ref{thm;my_init_shallow_main} with thm. \ref{thm;lecun_deep}, we observe that utilizing Initialization 1 requires quadratic overparameterization in the final hidden layer, as opposed to super quadratic overparameterization, for gradient descent to converge to a global minimum.   
\end{remark}

Many practitioners in machine learning often use the Uniform distribution to sample the weight matrices of a neural network at initialization. Motivated by this we define a second uniform initialization scheme as follows.

Fix a deep neural field $F$ with $L$ layers We define the following uniform initialization.
\begin{align}
& \text{ \textbf{Initialization 2} } \nonumber \\
    &(W^0_l)_{ij} \sim 
    \mathcal{U}(-1/\sqrt{n_{l-1}}, 1/\sqrt{n_{l-1}})\text{ for } l \neq L. \label{eqn;my_uinit_deep_a} \\
    &(W^0_L)_{ij} \sim 
    \mathcal{U}(-1/(n_{l-1}^{3/4}), 
    1/(n_{l-1}^{3/4}))\label{eqn;my_uinit_deep_b}
\end{align}
where $\mathcal{U}(a,b)$ denotes the Uniform distribution on the interval $[a, b]$.
The biases can be initialized to be $0$ or a very small number. In sec. 1 of the supp. material we give another way to initialize the biases.

\begin{remark}\label{rmk;siren_init}
    In \cite{siren}, Sitzmann et al. presented a uniform initialization for networks using a sine activation. In sec. 1 of the supp. material, we demonstrate the combination of our initialization \eqref{eqn;my_uinit_deep_a}, \eqref{eqn;my_uinit_deep_b} with theirs and provide comparisons in sec. 2 of the supp. material.
\end{remark}


\section{Experiments: Applications to Neural Fields}\label{sec;exps}



In this section, we empirically test the theory developed in sec. \ref{sec;theory_main}, focusing exclusively on the widely used initializations LeCun normal \cite{lecun2002efficient}, Xavier normal \cite{glorot2010understanding}, Kaiming normal \cite{he2015delving}, and their uniform counterparts. If unfamiliar with these initializations, please refer to sec. 2 of the supp. material. Additionally, we employ four activations: sinc \cite{ramasinghe2023effectiveness}, Gaussian \cite{ramasinghe22}, Gabor wavelet \cite{saragadam2023wire}, 
each with a frequency parameter $\omega$ (or $1/\omega^2$ for the Gaussian)
and
ReLU-PE \cite{tancik2020fourier} with a positional embedding layer. For details on tuning this hyperparameter and the frequencies used in each experiment, consult sec. 2 of the supp. material. Further experiments can also be found in sec. 2 of the supp. material.


\subsection{Practical Validation of the Theoretical Analysis}\label{sec;testing_scaling_laws}

We conduct empirical testing to validate the theories presented in sections \ref{sec;shallow_nets} and \ref{sec;deep_nets}. In section \ref{sec;shallow_nets}, we derived theorem \ref{thm;Lecun_shallow}, demonstrating that a shallow feedforward neural network utilizing activation functions such as sine, sinc, Gaussian, or wavelet, and initialized according to LeCun, Xavier, or Kaiming initialization, necessitates superlinear growth in the width of the hidden layer as the dataset size increases for gradient descent to converge to a global minimum.

It is noteworthy that this growth requirement is less stringent than what is established for ReLU activation (with or without PE) in Ngyuen's work \cite{nguyen2021proof}, which asserts a quadratic growth in the number of data samples. Consequently, we anticipate observing that a ReLU (with or without PE) network demands more parameters than its sinc-activated counterpart when trained on the same dataset until convergence.

\paragraph{Shallow Experiment:} In our investigation, we performed a 1-dimensional curve fitting experiment on the function $f(x) = \sin(2\pi x) + \sin(6\pi x) + \sin(10\pi x)$. We systematically sampled the curve at intervals of $10$, $20$, $50$, $75$, $100$, $125$, $150$, $175$, and $200$ points, creating nine datasets of varying sizes for our training data.

Subsequently, we employed three networks with sinc activation and one hidden layer, along with three networks featuring ReLU-PE activation and one hidden layer. The positional embedding layer had dimension $8$ and employed a Random Fourier Feature (RFF) type embedding \cite{tancik2020fourier}.
Each dataset underwent training using full batch gradient descent until reaching a PSNR value of 35dB. We increased the number of parameters with the growth in dataset size, adjusting them until the respective networks achieved convergence at the target PSNR.

\begin{figure}[t]
    \centering
    \includegraphics[width=1.\linewidth]
    {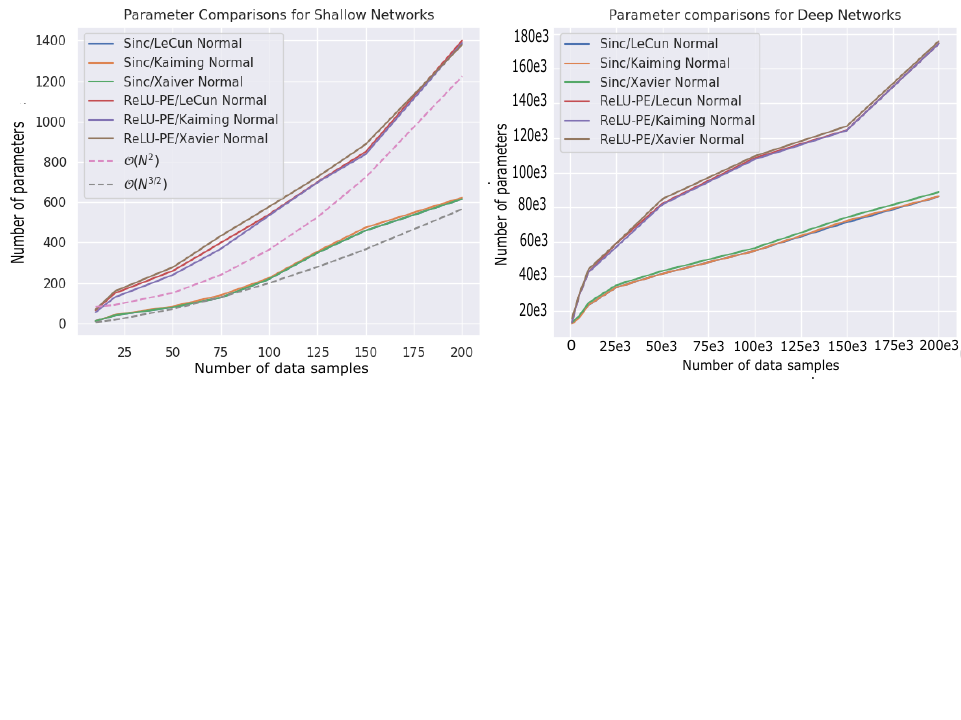}
    \vspace{-10em}
    \caption{Comparing how many parameters are needed for a ReLU-PE and sinc network to converge with different intializations and data set sizes. Left figure shows results for shallow networks on a 1-dim. curve fitting task. Right figure shows results for deep networks on a image regression task. For all initializations, the sinc activated networks require much less parameters to converge than the ReLU-PE ones.}
    \label{fig:relu_sinc_compare}
\end{figure}

Initialization for both sinc and ReLU-PE networks was performed using LeCun, Xavier, and Kaiming initializations. The results of this experiment are illustrated in fig. \ref{fig:relu_sinc_compare} (left). As anticipated by thm. \ref{thm;Lecun_shallow}, the sinc-activated networks exhibited a significantly lower parameter requirement for convergence as the dataset size increased. Furthermore, as fig. \ref{fig:relu_sinc_compare} (left) shows the 
sinc networks had parameter growth similar to $\mathcal{O}(N^{3/2})$, as predicited by thm. \ref{thm;Lecun_shallow}. The
ReLU-PE networks had parameter growth as $\mathcal{O}(N^2)$ as predicted in \cite{nguyen2021proof}.

\paragraph{Deep Experiment:} For the case of deep networks we ran a similar experiment to the above shallow networks except that this time we used an image regression task.

The task was to reconstruct a $512 \times 512$ Lion image. 
Given pixel coordinates $\mathbf{x} \in \mathbb{R}^2$, the aim of the task is to a network $\mathcal{N}$ to regress the associated RGB values $\mathbf{c} \in \mathbb{R}^3$~\cite{siren, ramasinghe22}.
We sampled $1000$, $5000$, $10000$, $25000$, $50000$, $100000$, $150000$ and $200000$ pixel coordinates, creating a total of $8$ datasets of varying size as the training data.

We employed three networks with sinc activation and four hidden layers, along with three networks featuring ReLU-PE activation and four hidden layers. The PE-layer was $16$ dimensional and used Random Fourier Features (RFF) as the positional embedding technique \cite{tancik2020fourier}. The first three hidden layers of all networks had $64$ neurons each. We increased the number of parameters by only increasing the width of the last hidden layer as this was shown to suffice from thm. \ref{thm;lecun_deep}. 


Each dataset underwent training using full batch gradient descent until reaching a PSNR value of 25dB. Initialization for both sinc and ReLU-PE networks was performed using LeCun, Xavier, and Kaiming initializations. The results of this experiment are illustrated in fig. \ref{fig:relu_sinc_compare} (right).
As predicted by thm. \ref{thm;Lecun_shallow}, the sinc-activated networks exhibited a significantly lower parameter requirement for convergence as the dataset size increased. Nevertheless, it is essential to highlight that beyond a dataset size of $30,000$, the sinc networks achieved convergence using fewer parameters than the actual number of samples. This discrepancy arises from our decision to set the cutoff point at 25dB. The rationale behind this choice was rooted in practical considerations related to memory constraints. Given that we employed full-batch training, higher dB values necessitated a substantially greater number of parameters, leading to memory-related issues. 

\begin{figure}[t]
    \centering
    \includegraphics[width=1.\linewidth]
    {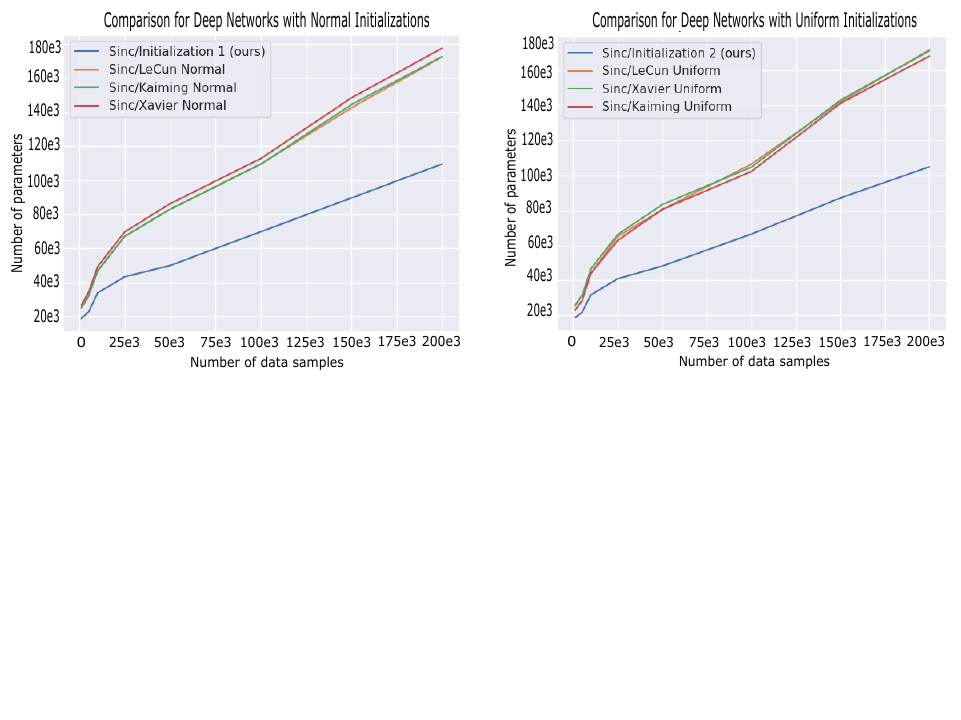}
    \vspace{-10em}
    \caption{Comparing the performance of deep networks with sinc activation across image regression tasks, utilizing four distinct initialization schemes. Networks were trained until reaching a 25dB PSNR. On the left, we observe the outcomes with four normal initializations, showcasing that our initialization demands the fewest parameters for convergence. On the right, the comparison extends to four different uniform initializations, where our approach emerges as the most effective.}
    \label{fig:my_init_compare_stds}
\end{figure}

The second experiment was to test thm. \ref{thm;my_init_deep_main}. In order to do this we carried out a similar experiment to the above deep network experiment, using the same datasets and data instance. In this setting 
we considered four sinc networks, each initialized with LeCun normal, Xavier normal, Kaiming normal and \textbf{initialization 1} (see \eqref{eqn;my_init_deep_a}). We trained each each network until it reached a PSNR of 25dB, increasing the width of the final hidden layer as the dataset size increased. Fig. \ref{fig:my_init_compare_stds} (left) shows the results of the experiment. As predicted by thm. \ref{thm;my_init_deep_main}, the network employing our initialization 1 needs far less parameters to converge. We repeated the experiments this time initializing the networks with the uniform distributions, \textbf{initialization 2} (see \eqref{eqn;my_init_deep_b}), LeCun uniform, Xavier uniform, Kaiming uniform. Fig. \ref{fig:my_init_compare_stds} (right) shows that the network initialized with initialization 2 required less parameters to converge than all other networks.

\subsection{Single Image Super Resolution}\label{sec;ssir}

We test our initializations on an image super resolution task. 
We take the approach considered in \cite{saragadam2023wire}, where a $4 \times$ super resolution is conducted on an image from the DIV2K dataset \cite{agustsson2017ntire, timofte2017ntire}. The problem is cast as solving $y = Ax$, where the operator $A$ implements a $4 \times$ downsampling (with no aliasing). We then solve for $x$ as the output of a neural field. 

\begin{figure}[t]
    \centering
    \includegraphics[width=.8\linewidth]
    {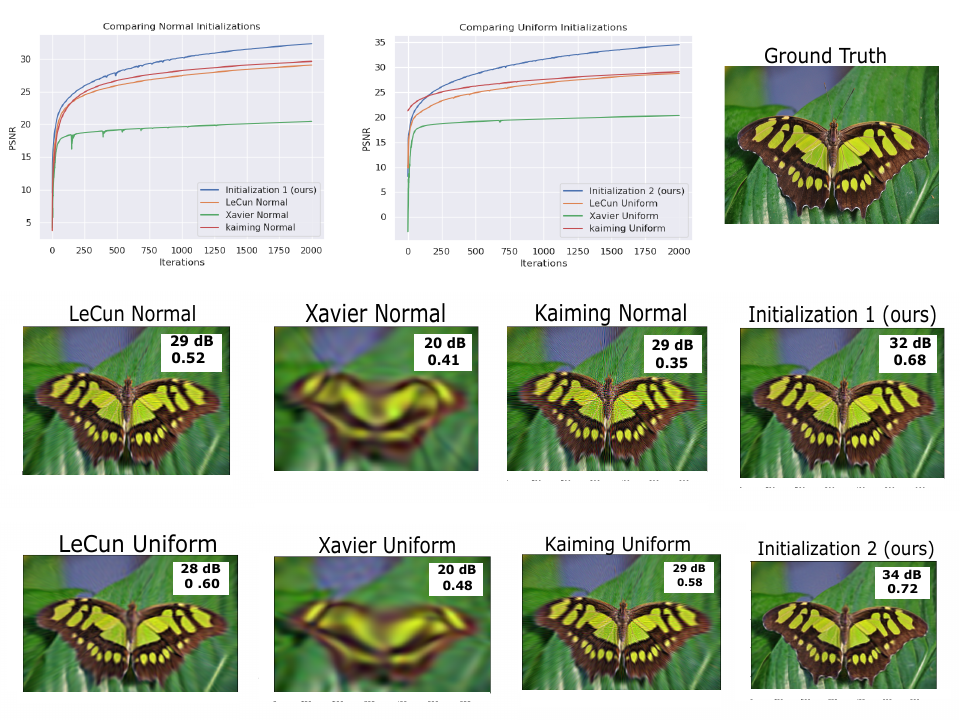}
    \caption{The figure shows the results for a $4 \times$ single image super resolution with four normal initializations and four uniform initializations. Networks initialized with initialization 1 (our) and initialization 2 (our) produced the highest train dB and SSIM at convergence. Zoom in for better viewing.}
    \label{fig:ssir}
\end{figure}

We explored the impact of four normal initializations and four uniform initializations on a Gaussian-activated network with two hidden layers. The Gaussian activation was characterized by a variance of $0.1^2$, a choice validated as optimal across all initializations and commonly adopted by practitioners \cite{ramasinghe2023effectiveness, chng2022garf, ramasinghe22, saratchandran2023curvature}. Subsequently, all networks underwent training using the Adam optimizer.
Figure \ref{fig:ssir} presents the outcomes of this investigation. Notably, among the normal initializations, our first initialization demonstrated superior performance, while among the uniform initializations, our second initialization excelled. In each case, Structural Similarity (SSIM \cite{wang2004image}.) was computed, with both our first and second initializations consistently producing the highest SSIM values.

\subsection{Occupancy Fields}\label{sec;occupancy}
\begin{figure}[t]
    \centering
    \includegraphics[width=1.\linewidth]
    {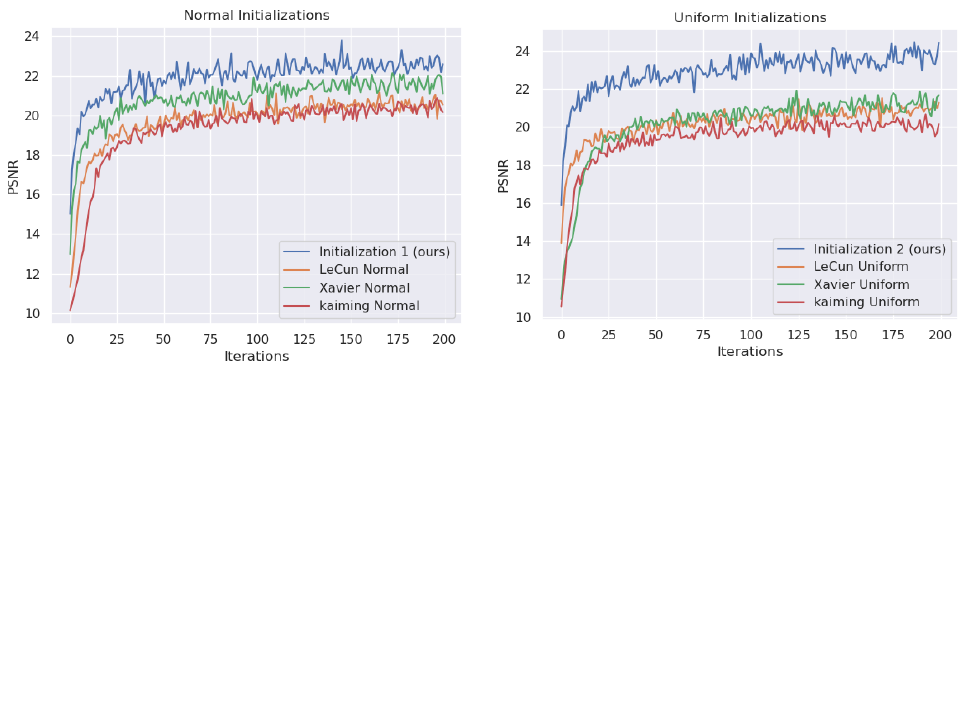}
    \vspace{-11em}
    \caption{Top left; comparison of normal initializations. Top right; comparison of uniform initializations. In both cases our initialization performs better reaching a higher PSNR. Bottom summary of final train PSNR and IOU accuracy.}
    \label{fig:occupancy}
\end{figure}

\vspace{-0.5em}
We optimize a binary occupancy field, which represents a 3D shape as the decision boundary of a neural network ~\cite{wang2021spline, gropp2020implicit}. We use the \textit{thai statue} instance obtained from XYZ RGB Inc. 
We trained two groups of four networks, each activated with the Gabor wavelet \cite{saragadam2023wire}. The first group of four networks were trained with four normal initializations, Lecun normal, Xavier normal, Kaiming normal and initialization 1 
(see \eqref{eqn;my_init_deep_a}, \eqref{eqn;my_init_deep_b}). The second group of four networks were trained with LeCun uniform, Xavier uniform, Kaiming Uniform and initialization 2 (see \eqref{eqn;my_uinit_deep_a}, \eqref{eqn;my_uinit_deep_b}). 
All networks were trained with the Adam optimizer. For accuracy testing we used the performance metric given by Intersection Over Union (IOU) .
Fig. \ref{fig:occupancy} and table \ref{table:occupancy_table_results} details the results of the experiments. As can be seen from the figure, both our initializations converge to a higher PSNR, $2-4$dB higher than the others, and have the highest IOU. Figures of reconstructed meshes can be found in the sec. 2 of supp. material.

\begin{table}
\centering
\begin{tabular}{ |c|c|c| } 
 \hline
 {} & \textbf{Train PSNR} (dB) & \textbf{IOU} \\ 
 \hline
Initialization 1 (ours) & \textbf{22.7} & \textbf{0.89} \\ 
 \hline
 LeCun Normal & 20.3 & 0.81 \\ 
 \hline
 Xavier Normal & 21.2 & 0.84 \\ 
 \hline
 Kaiming Normal & 19.9 & 0.80 \\ 
 \hline
 Initialization 2 (ours) & \textbf{24.5} & \textbf{0.91} \\ 
 \hline
 LeCun Uniform  & 21.2 & 0.86 \\ 
 \hline
 Xaiver Uniform & 21.6 & 0.87 \\ 
 \hline
 Kaiming Uniform & 20.2 & 0.82 \\ 
 \hline
\end{tabular}%
\caption{Table showing results of each initialization on an Occupancy task.}
\label{table:occupancy_table_results}
\end{table}

\subsection{Neural Radiance Fields (NeRF)}\label{sec;nerf}
\vspace{-0.5em}
\begin{figure}[t]
    \centering
    \includegraphics[width=1.\linewidth]
    {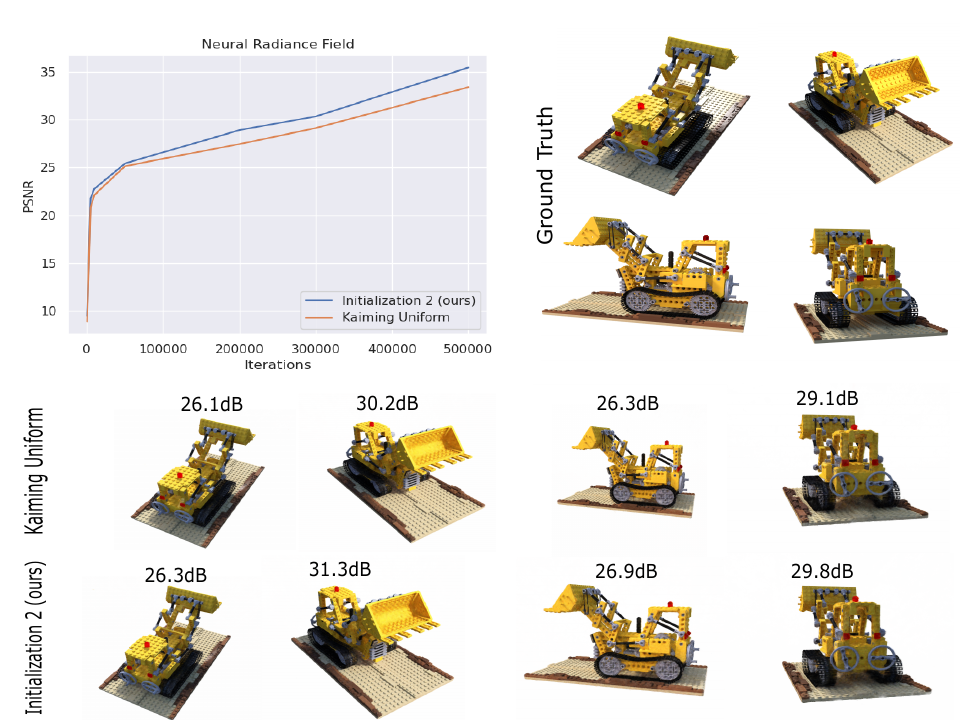}
    \caption{Training results comparison for two Gaussian-activated NeRFs: one with Kaiming uniform initialization and the other with Initialization 2 (see \eqref{eqn;my_uinit_deep_a} and \eqref{eqn;my_uinit_deep_b}). Top-left: Training PSNR. Top-right: Four ground truth instances for testing. Bottom: Test PSNRs. Initialization 2 consistently outperforms the Kaiming uniform-initialized NeRF in both training and test accuracy.}
    \label{fig:nerf}
\end{figure}

NeRF has recently emerged as a compelling method, leveraging a Multi-Layer Perceptron (MLP) to model 3D objects and scenes based on multi-view 2D images. This innovative approach exhibits promise in achieving high-fidelity reconstructions for novel view synthesis tasks \cite{nerf, kilonerf, chng2022garf, lin2021barf}. Given 3D points $\mathbf{x} \in \mathbb{R}^3$ and viewing directions, NeRF is designed to estimate the radiance field of a 3D scene. This field maps each input 3D coordinate to its corresponding volume density $\sigma \in \mathbb{R}$ and directional emitted color $\mathbf{c} \in \mathbb{R}^3$ \cite{nerf, lin2021barf, chng2022garf}.

NeRF is commonly trained using Kaiming uniform initialization for optimal outcomes. Our experiment involved two Gaussian-activated NeRFs: one initialized with Kaiming uniform and the other with Initialization 2 (see \eqref{eqn;my_uinit_deep_a}, \eqref{eqn;my_uinit_deep_b}). 
We used the Lego instance from the NeRF real synthetic data set. All networks were trained with the Adam optimizer. 
Fig. \ref{fig:nerf} displays the results, indicating that Initialization 2 achieved a higher training PSNR by 1.1dB. Testing across 24 unseen views revealed that Initialization 2 consistently outperformed Kaiming initialization, with a test difference ranging from 0.1 to 1.1dB across different scenes.
For more details please see sec. 2 of the supp. material.

\subsection{Physics Informed Neural Networks (PINNs)}\label{sec;fluids}
\vspace{-0.5em}
Physics informed neural networks are an innovative neural architecture that parameterize a physics field arising as the solution of a partial differential equation (PDE). 
For an introduction to PINNs we refer the reader to \cite{raissi2019physics}.

We consider the 2D incompressible Navier-Stokes equations as considered in \cite{raissi2019physics}. 


\begin{align}
    u_t + uu_x + 0.01u_y &= -p_x + 0.01(u_{xx} + u_{yy})\label{eqn;navier_eqn_1} \\
    v_t + uv_x + 0.01v_y &= -p_y + 0.01(v_{xx} + v_{yy})\label{eqn;navier_eqn_2}
\end{align}
where $u(x,y,t)$ denotes the $x$-component of the velocity field of the fluid, and 
$v(x,y,t)$ denotes the $y$-component of the velocity field. The term $p(t,x,y)$ is the pressure. The domain of the problem is 
$[-15, 25]\times [-8, 8] \times [0,20]$.

The fluid field solution to equations \eqref{eqn;navier_eqn_1} and \eqref{eqn;navier_eqn_2} can be parameterized by a neural field that is a PINN. The two PDE equations \eqref{eqn;navier_eqn_1} and \eqref{eqn;navier_eqn_2} are embedded into the loss function enforcing a physical constraint on the network as it trains. 

We trained eight Gaussian-activated PINNs with three hidden layers and a width of 128. Four PINNs used LeCun, Xavier, Kaiming, and Initialization 1 (\eqref{eqn;my_init_deep_a}, \eqref{eqn;my_init_deep_b}) for normal initialization, while the other four used corresponding uniform initializations. Training employed an Adam optimizer with full batch. For detailed training setup, please refer to sec. 3 in the supplementary material.

Fig. \eqref{fig:navier} shows the training PSNR's of each of the networks. As can be seen from the figure, the PINNs initialized with initialization 1 and 2 reach a higher dB in total loss and PDE loss.

\begin{figure}[t]
    \centering
    \includegraphics[width=.9\linewidth]
    {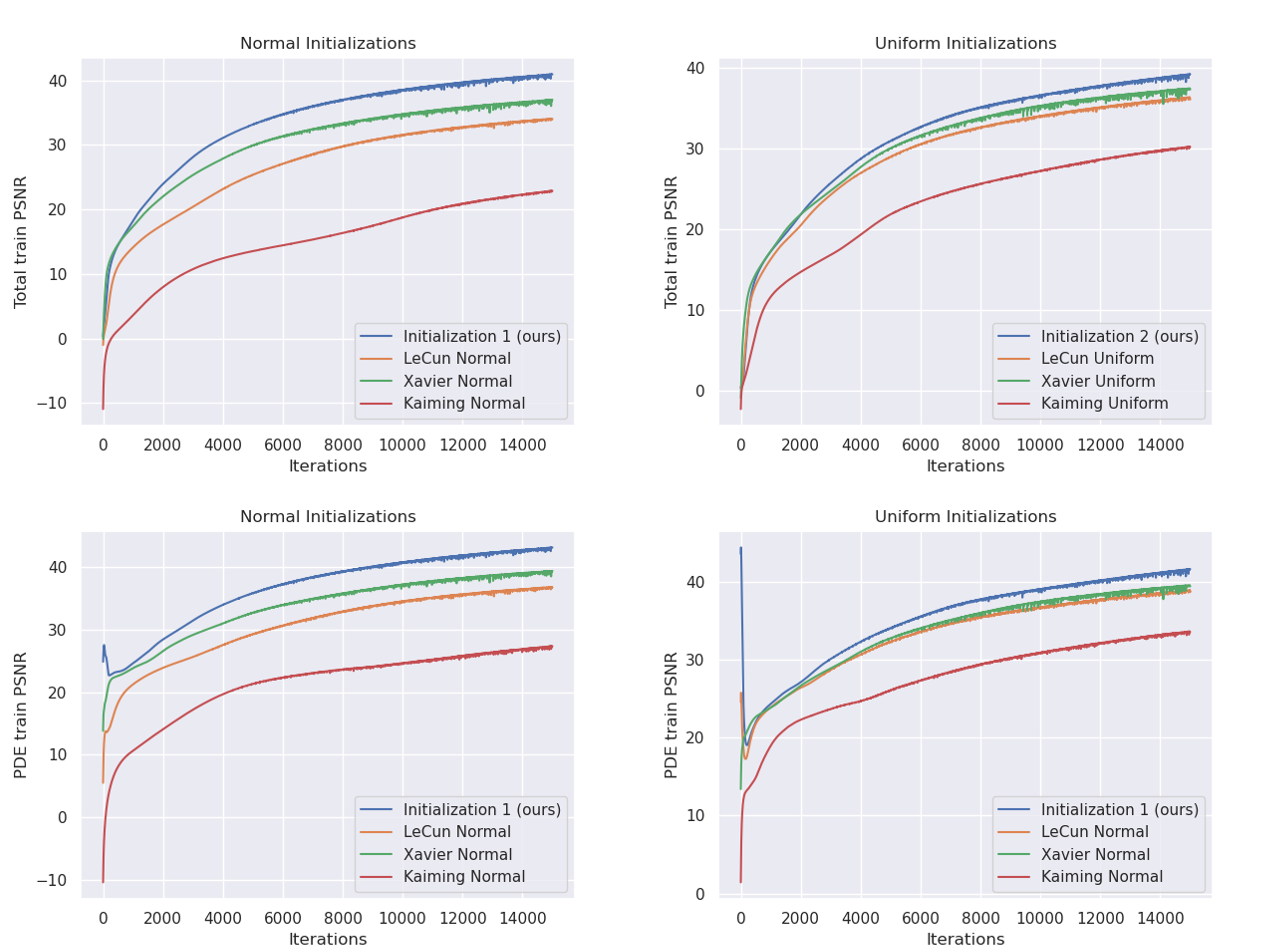}
    \caption{Results from training eight different Gaussian activated PINNs, each with a different initialization. Top row plots the total loss PSNR (MSE loss + PDE loss) and bottom row plots PDE loss PSNR. In both cases the PINNs initialized with our initialization reach a higher dB in both total and PDE loss.}
    \label{fig:navier}
\end{figure}

\section{Conclusion}\label{sec;conclusion}
\vspace{-0.5em}
This paper established a theoretical framework for the optimal scaling of neural fields as dataset sizes expand, ensuring optimal convergence during gradient descent. We uncovered a link between this scalability challenge and the activation and initialization of the network. Our theoretical framework yielded state-of-the-art results for both shallow and deep networks. Moreover, leveraging our theoretical insights, we proposed a novel initialization scheme and validated its efficacy across diverse neural field applications.


\section{Limitations}\label{sec;limitations}
\vspace{-0.5em}

This paper delves into the theory to identify the necessary degree of overparameterization for gradient descent to reach a global minimum. For practical scenarios, training typically uses mini-batches. Unfortunately, our theoretical results do not yet apply to mini-batch training and we suggest that future research applying our findings to mini-batch scenarios could provide useful methodologies on how to scale networks for such training.

\section{Acknowledgements}
Simon Lucey acknowledges support from the Australian Research Council (ARC) through the Discovery Project DP220103803.


{
    \small
    \bibliographystyle{ieeenat_fullname}
    \bibliography{main}

@String(ICCV= {Int. Conf. Comput. Vis.})

@String(ECCV= {Eur. Conf. Comput. Vis.})

@String(NIPS= {Adv. Neural Inform. Process. Syst.})

@String(ICCV  = {ICCV})

@String(ECCV  = {ECCV})

@String(NIPS  = {NeurIPS})

@article{davidson2001local,
  title={Local operator theory, random matrices and Banach spaces},
  author={Davidson, Kenneth R and Szarek, Stanislaw J},
  journal={Handbook of the geometry of Banach spaces},
  volume={1},
  number={317-366},
  pages={131},
  year={2001},
  publisher={North-Holland, Amsterdam}
}

@article{sitzmann2020implicit,
  title={Implicit neural representations with periodic activation functions},
  author={Sitzmann, Vincent and Martel, Julien and Bergman, Alexander and Lindell, David and Wetzstein, Gordon},
  journal={Advances in neural information processing systems},
  volume={33},
  pages={7462--7473},
  year={2020}
}

@inproceedings{nguyen2021proof,
  title={On the proof of global convergence of gradient descent for deep relu networks with linear widths},
  author={Nguyen, Quynh},
  booktitle={International Conference on Machine Learning},
  pages={8056--8062},
  year={2021},
  organization={PMLR}
}

@inproceedings{siren,
author = {V. Sitzmann and J. Martel and A. Bergman and D. Lindell and G. and Wetzstein},
title = {{Implicit Neural Representations with Periodic Activation Functions}},
booktitle = {NIPS},
year = 2020
}

@inproceedings{ramasinghe22,
author = {S. Ramasinghe and S. Lucey},
title = {{Beyond Periodicity: Towards a Unifying Framework for Activations in Coordinate-MLPs}},
booktitle = {ECCV},
year = 2022
}

@inproceedings{chng2022garf,
  title={Gaussian activated neural radiance fields for high fidelity reconstruction and pose estimation},
  author={Chng, Shin-Fang and Ramasinghe, Sameera and Sherrah, Jamie and Lucey, Simon},
  booktitle={Computer Vision--ECCV 2022: 17th European Conference, Tel Aviv, Israel, October 23--27, 2022, Proceedings, Part XXXIII},
  pages={264--280},
  year={2022},
  organization={Springer}
}

@article{saratchandran2023curvature,
  title={Curvature-Aware Training for Coordinate Networks},
  author={Saratchandran, Hemanth and Chng, Shin-Fang and Ramasinghe, Sameera and MacDonald, Lachlan and Lucey, Simon},
  journal={arXiv preprint arXiv:2305.08552},
  year={2023}
}

@inproceedings{saragadam2023wire,
  title={Wire: Wavelet implicit neural representations},
  author={Saragadam, Vishwanath and LeJeune, Daniel and Tan, Jasper and Balakrishnan, Guha and Veeraraghavan, Ashok and Baraniuk, Richard G},
  booktitle={Proceedings of the IEEE/CVF Conference on Computer Vision and Pattern Recognition},
  pages={18507--18516},
  year={2023}
}

@article{ramasinghe2023effectiveness,
  title={On the effectiveness of neural priors in modeling dynamical systems},
  author={Ramasinghe, Sameera and Saratchandran, Hemanth and Shevchenko, Violetta and Lucey, Simon},
  journal={arXiv preprint arXiv:2303.05728},
  year={2023}
}

@incollection{lecun2002efficient,
  title={Efficient backprop},
  author={LeCun, Yann and Bottou, L{\'e}on and Orr, Genevieve B and M{\"u}ller, Klaus-Robert},
  booktitle={Neural networks: Tricks of the trade},
  pages={9--50},
  year={2002},
  publisher={Springer}
}

@inproceedings{he2015delving,
  title={Delving deep into rectifiers: Surpassing human-level performance on imagenet classification},
  author={He, Kaiming and Zhang, Xiangyu and Ren, Shaoqing and Sun, Jian},
  booktitle={Proceedings of the IEEE international conference on computer vision},
  pages={1026--1034},
  year={2015}
}

@inproceedings{glorot2010understanding,
  title={Understanding the difficulty of training deep feedforward neural networks},
  author={Glorot, Xavier and Bengio, Yoshua},
  booktitle={Proceedings of the thirteenth international conference on artificial intelligence and statistics},
  pages={249--256},
  year={2010},
  organization={JMLR Workshop and Conference Proceedings}
}

@article{sun2021coil,
  title={Coil: Coordinate-based internal learning for imaging inverse problems},
  author={Sun, Yu and Liu, Jiaming and Xie, Mingyang and Wohlberg, Brendt and Kamilov, Ulugbek S},
  journal={arXiv preprint arXiv:2102.05181},
  year={2021}
}

@inproceedings{skorokhodov2021adversarial,
  title={Adversarial generation of continuous images},
  author={Skorokhodov, Ivan and Ignatyev, Savva and Elhoseiny, Mohamed},
  booktitle={Proceedings of the IEEE/CVF Conference on Computer Vision and Pattern Recognition},
  pages={10753--10764},
  year={2021}
}

@inproceedings{chen2021learning,
  title={Learning continuous image representation with local implicit image function},
  author={Chen, Yinbo and Liu, Sifei and Wang, Xiaolong},
  booktitle={Proceedings of the IEEE/CVF conference on computer vision and pattern recognition},
  pages={8628--8638},
  year={2021}
}

@article{sitzmann2019scene,
  title={Scene representation networks: Continuous 3d-structure-aware neural scene representations},
  author={Sitzmann, Vincent and Zollh{\"o}fer, Michael and Wetzstein, Gordon},
  journal={Advances in Neural Information Processing Systems},
  volume={32},
  year={2019}
}

@inproceedings{li20223d,
  title={3d neural scene representations for visuomotor control},
  author={Li, Yunzhu and Li, Shuang and Sitzmann, Vincent and Agrawal, Pulkit and Torralba, Antonio},
  booktitle={Conference on Robot Learning},
  pages={112--123},
  year={2022},
  organization={PMLR}
}

@article{chen2022fully,
  title={Fully body visual self-modeling of robot morphologies},
  author={Chen, Boyuan and Kwiatkowski, Robert and Vondrick, Carl and Lipson, Hod},
  journal={Science Robotics},
  volume={7},
  number={68},
  pages={eabn1944},
  year={2022},
  publisher={American Association for the Advancement of Science}
}

@article{tancik2020fourier,
  title={Fourier features let networks learn high frequency functions in low dimensional domains},
  author={Tancik, Matthew and Srinivasan, Pratul and Mildenhall, Ben and Fridovich-Keil, Sara and Raghavan, Nithin and Singhal, Utkarsh and Ramamoorthi, Ravi and Barron, Jonathan and Ng, Ren},
  journal={Advances in Neural Information Processing Systems},
  volume={33},
  pages={7537--7547},
  year={2020}
}

@article{nerf,
  title={Nerf: Representing scenes as neural radiance fields for view synthesis},
  author={Mildenhall, Ben and Srinivasan, Pratul P and Tancik, Matthew and Barron, Jonathan T and Ramamoorthi, Ravi and Ng, Ren},
  journal={Communications of the ACM},
  volume={65},
  number={1},
  pages={99--106},
  year={2021},
  publisher={ACM New York, NY, USA}
}

@inproceedings{agustsson2017ntire,
  title={Ntire 2017 challenge on single image super-resolution: Dataset and study},
  author={Agustsson, Eirikur and Timofte, Radu},
  booktitle={Proceedings of the IEEE conference on computer vision and pattern recognition workshops},
  pages={126--135},
  year={2017}
}

@inproceedings{timofte2017ntire,
  title={Ntire 2017 challenge on single image super-resolution: Methods and results},
  author={Timofte, Radu and Agustsson, Eirikur and Van Gool, Luc and Yang, Ming-Hsuan and Zhang, Lei},
  booktitle={Proceedings of the IEEE conference on computer vision and pattern recognition workshops},
  pages={114--125},
  year={2017}
}

@article{wang2004image,
  title={Image quality assessment: from error visibility to structural similarity},
  author={Wang, Zhou and Bovik, Alan C and Sheikh, Hamid R and Simoncelli, Eero P},
  journal={IEEE transactions on image processing},
  volume={13},
  number={4},
  pages={600--612},
  year={2004},
  publisher={IEEE}
}

@article{wang2021spline,
  title={Spline positional encoding for learning 3d implicit signed distance fields},
  author={Wang, Peng-Shuai and Liu, Yang and Yang, Yu-Qi and Tong, Xin},
  journal={arXiv preprint arXiv:2106.01553},
  year={2021}
}

@article{gropp2020implicit,
  title={Implicit geometric regularization for learning shapes},
  author={Gropp, Amos and Yariv, Lior and Haim, Niv and Atzmon, Matan and Lipman, Yaron},
  journal={ICML},
  year={2020}
}

@inproceedings{kilonerf,
author = {C. Reiser and S. Peng and Y. Liao and A. Geiger},
title = {{KiloNeRF: Speeding Up Neural Radiance Fields With Thousands of Tiny MLPs}},
booktitle = {ICCV},
year = 2021
}

@inproceedings{lin2021barf,
  title={Barf: Bundle-adjusting neural radiance fields},
  author={Lin, Chen-Hsuan and Ma, Wei-Chiu and Torralba, Antonio and Lucey, Simon},
  booktitle={Proceedings of the IEEE/CVF International Conference on Computer Vision},
  pages={5741--5751},
  year={2021}
}

@article{cooper2018loss,
  title={The loss landscape of overparameterized neural networks},
  author={Cooper, Yaim},
  journal={arXiv preprint arXiv:1804.10200},
  year={2018}
}

@article{zou2019improved,
  title={An improved analysis of training over-parameterized deep neural networks},
  author={Zou, Difan and Gu, Quanquan},
  journal={Advances in neural information processing systems},
  volume={32},
  year={2019}
}

@article{allen2019learning,
  title={Learning and generalization in overparameterized neural networks, going beyond two layers},
  author={Allen-Zhu, Zeyuan and Li, Yuanzhi and Liang, Yingyu},
  journal={Advances in neural information processing systems},
  volume={32},
  year={2019}
}

@article{oymak2020toward,
  title={Toward moderate overparameterization: Global convergence guarantees for training shallow neural networks},
  author={Oymak, Samet and Soltanolkotabi, Mahdi},
  journal={IEEE Journal on Selected Areas in Information Theory},
  volume={1},
  number={1},
  pages={84--105},
  year={2020},
  publisher={IEEE}
}

@inproceedings{du2019gradient,
  title={Gradient descent finds global minima of deep neural networks},
  author={Du, Simon and Lee, Jason and Li, Haochuan and Wang, Liwei and Zhai, Xiyu},
  booktitle={International conference on machine learning},
  pages={1675--1685},
  year={2019},
  organization={PMLR}
}

@inproceedings{arora2019fine,
  title={Fine-grained analysis of optimization and generalization for overparameterized two-layer neural networks},
  author={Arora, Sanjeev and Du, Simon and Hu, Wei and Li, Zhiyuan and Wang, Ruosong},
  booktitle={International Conference on Machine Learning},
  pages={322--332},
  year={2019},
  organization={PMLR}
}

@inproceedings{huang2020dynamics,
  title={Dynamics of deep neural networks and neural tangent hierarchy},
  author={Huang, Jiaoyang and Yau, Horng-Tzer},
  booktitle={International conference on machine learning},
  pages={4542--4551},
  year={2020},
  organization={PMLR}
}

@article{jacot2018neural,
  title={Neural tangent kernel: Convergence and generalization in neural networks},
  author={Jacot, Arthur and Gabriel, Franck and Hongler, Cl{\'e}ment},
  journal={Advances in neural information processing systems},
  volume={31},
  year={2018}
}

@inproceedings{allen2019convergence,
  title={A convergence theory for deep learning via over-parameterization},
  author={Allen-Zhu, Zeyuan and Li, Yuanzhi and Song, Zhao},
  booktitle={International Conference on Machine Learning},
  pages={242--252},
  year={2019},
  organization={PMLR}
}

@article{raissi2019physics,
  title={Physics-informed neural networks: A deep learning framework for solving forward and inverse problems involving nonlinear partial differential equations},
  author={Raissi, Maziar and Perdikaris, Paris and Karniadakis, George E},
  journal={Journal of Computational physics},
  volume={378},
  pages={686--707},
  year={2019},
  publisher={Elsevier}
}
}



\end{document}